\documentstyle[colacl,epsf]{article}

\title{Bunsetsu Identification Using Category-Exclusive Rules}

\author{\vspace*{2mm}Masaki Murata \hspace{0.6cm} Kiyotaka Uchimoto \hspace{0.6cm} Qing Ma \hspace{0.6cm} Hitoshi Isahara\\
           Communications Research Laboratory, 
           Ministry of Posts and Telecommunications\\
           588-2, Iwaoka, Nishi-ku, Kobe, 651-2492, Japan\\
           tel:+81-78-969-2181 \, fax:+81-78-969-2189 \, http://www-karc.crl.go.jp/ips/murata\\
           \{murata,uchimoto,qma,isahara\}@crl.go.jp
}

\def\None#1{}

\def\Small#1{}

\def\OK#1{}

\def\Out#1{}

\begin{document}


\setlength{\footskip}{1cm}
\thispagestyle{plain}
\pagestyle{plain}
\maketitle
\begin{abstract}
This paper describes two new bunsetsu identification methods 
using supervised learning. 
Since Japanese syntactic analysis is 
usually done after bunsetsu identification, 
bunsetsu identification is important 
for analyzing Japanese sentences. 
In experiments 
comparing the four previously available machine-learning methods 
(decision tree, maximum-entropy method, example-based 
approach and decision list) 
and two new methods using category-exclusive rules, 
the new method using the category-exclusive rules 
with the highest similarity performed best. 
\end{abstract}

\section{Introduction}

This paper is about machine learning methods for identifying 
{\it bunsetsu}s, which correspond to English phrasal units 
such as noun phrases and prepositional phrases. 
Since Japanese syntactic analysis is 
usually done after bunsetsu identification \cite{uchimoto:eacl99}, 
identifying bunsetsu is important for analyzing Japanese sentences. 
The conventional studies on bunsetsu 
identification\footnote{Bunsetsu identification is a problem similar to 
chunking \cite{ramchaw95,sang99} in other languages. } 
have used hand-made rules \cite{kameda:nlpsympo95e,KNP2.0b6_e}, 
but bunsetsu identification is not an easy task.  
Conventional studies used many hand-made rules 
developed at the cost of many man-hours. 
Kurohashi, for example,  made 146 rules 
for bunsetsu identification \cite{KNP2.0b6_e}. 

In an attempt to reduce the number of man-hours, 
we used machine-learning methods for bunsetsu identification. 
Because it was not clear which machine-learning method 
would be the one most appropriate for bunsetsu identification, 
so we tried a variety of them. 
In this paper we report experiments 
comparing four machine-learning methods 
(decision tree, maximum entropy, example-based, and 
decision list methods) and our new methods 
using category-exclusive rules. 

\section{Bunsetsu identification problem}

\begin{figure*}[t]
\small
  \begin{center}
  \begin{tabular}{llllllll}
{\it boku} & {\it ga} & $|$  & {\it bunsetsu} &  {\it wo} &  $|$ &  {\it matomeageru} & . \\
(I) & {\footnotesize \sf nominative-case particle} & &
    (bunsetsu) & {\footnotesize \sf objective-case particle} &   & 
 (identify) & . \\
    \multicolumn{8}{l}{(I identify bunsetsu.)} \\
  \end{tabular}
\end{center}
\caption{Example of identified bunsetsus}
    \label{fig:ex_sent}
\end{figure*}

\begin{figure*}[t]
\small
  \begin{center}
\begin{tabular}[h]{|lcllll|}\hline
     &        & {\it bun} & {\it wo} & {\it kugiru} & . \\
     &        & (sentence) & ({\sf obj}) & (divide) & . \\
     &        & \multicolumn{4}{l|}{((I) divide sentences)}\\
Major POS & & Noun & Particle & Verb & Symbol\\[-0cm]
Minor POS & & Normal Noun & Case-Particle & Normal Form & Punctuation\\[-0cm]
Semantics &  & $\times$ & None & 217 & $\times$\\[-0cm]
Word&  & $\times$ & {\it wo} & {\it kugiru} & $\times$\\\hline
\end{tabular}
    \caption{Information used in bunsetsu identification}
    \label{fig:kaiseki_info}
  \end{center}
\end{figure*}

\begin{figure*}[t]
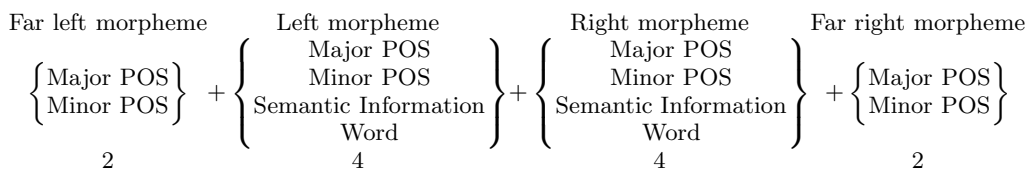

\small
  \begin{center}
{
\[
\begin{array}[h]{c@{}c@{}c@{}c}
\mbox{Far left morpheme} & \mbox{Left morpheme}  & \mbox{Right morpheme} & \mbox{Far right morpheme}\\
\left\{
\begin{array}[h]{@{}c@{}}
\mbox{Major POS}  \\
\mbox{Minor POS}  
\end{array}
\right\} 
&
+\left\{
\begin{array}[h]{@{}c@{}}
\mbox{Major POS}  \\
\mbox{Minor POS}  \\
\mbox{Semantic Information}  \\
\mbox{Word}  
\end{array}
\right\} 
&
+\left\{
\begin{array}[h]{@{}c@{}}
\mbox{Major POS}  \\
\mbox{Minor POS}  \\
\mbox{Semantic Information}  \\
\mbox{Word}  
\end{array}
\right\} 
&
+\left\{
\begin{array}[h]{@{}c@{}}
\mbox{Major POS}  \\
\mbox{Minor POS}
\end{array}
\right\} \\
2 & 4 & 4 & 2
\end{array}
\]
}
\vspace*{-0.5cm}
    \caption{Features used in the decision-tree method}
    \label{fig:info_for_dt}
  \end{center}
\end{figure*}

We conducted experiments on the following 
supervised learning methods for identifying bunsetsu: 
%
\begin{itemize}
\item 
  Decision tree method 
\item 
  Maximum entropy method
\item 
  Example-based method (use of similarity)
\item 
  Decision list (use of probability and frequency)
\item 
  Method 1 (use of exclusive rules)
\item 
  Method 2 (use of exclusive rules with the highest similarity). 
\end{itemize}

In general, 
bunsetsu identification is done after morphological 
and before syntactic analysis. 
Morphological analysis corresponds to part-of-speech tagging in English. 
Japanese syntactic structures are usually 
represented by the relations between bunsetsus, 
which correspond to phrasal units such as 
a noun phrase or a prepositional phrase in English. 
So, bunsetsu identification is important in Japanese sentence analysis. 

In this paper, 
we identify a bunsetsu by using information 
from a morphological analysis. 
Bunsetsu identification is treated as 
the task of deciding whether to 
insert a ``$|$'' {\it mark} to indicate 
the partition between two bunsetsus 
as in Figure \ref{fig:ex_sent}. 
Therefore, bunsetsu identification is done 
by judging whether a partition mark should be inserted 
between two adjacent morphemes or not. 
(We do not use 
the inserted partition mark in the following analysis 
in this paper for the sake of simplicity.) 

Our bunsetsu identification method uses 
the morphological information of 
the two preceding and 
two succeeding morphemes of 
an analyzed space between two adjacent morphemes. 
We use the following morphological information: 
\begin{enumerate}
\item[(i)] 
Major part-of-speech (POS) category,\footnote{
Part-of-speech categories follow those of JUMAN \cite{JUMAN3.5_e}.}
\item[(ii)] 
Minor POS category or inflection type,  
\item[(iii)] 
Semantic information (the first three-digit number of 
a category number as used in ``BGH'' \cite{NLRI64ae}), 
\item[(iv)] 
Word (lexical information). 
\end{enumerate}
For simplicity we do not use 
the ``Semantic information'' and ``Word'' 
in either of the two outside morphemes. 

Figure \ref{fig:kaiseki_info} shows the information 
used to judge whether or not to insert 
a partition mark in the space between two adjacent morphemes, 
``{\it wo} ({\sf obj})'' and ``{\it kugiru} (divide),'' 
in the sentence ``{\it bun wo kugiru. } ((I) divide sentences).'' 

\section{Bunsetsu identification process for each machine-learning method}
\label{sec:shuhou}

\subsection{Decision-tree method}

In this work 
we used the program C4.5 \cite{c4.5} 
for the decision-tree learning method. 
The four types of information, 
(i) major POS, (ii) minor POS, (iii) semantic information, and 
(iv) word, mentioned in the previous section were also used 
as features with the decision-tree learning method. 
As shown in Figure \ref{fig:info_for_dt}, 
the number of features is 12 ($2 + 4 + 4 + 2$) 
because we do not use (iii) semantic information and (iv) word 
information from the two outside morphemes. 

In Figure \ref{fig:kaiseki_info}, 
for example, 
the value of the feature `the major POS of the far left morpheme' is 
`Noun.' 

\subsection{Maximum-entropy method}

The maximum-entropy method is useful with sparse data conditions 
and has been used by many researchers 
\cite{berger:cl96,ratnaparkhi:emnlp96,ratnaparkhi:emnlp97,borthwick:96,uchimoto:eacl99}. 
In our maximum-entropy experiment 
we used Ristad's system \cite{ristad98}. 
The analysis is performed by 
calculating the probability of inserting 
or not inserting a partition mark, 
from the output of the system. 
Whichever probability is higher is selected as the desired answer. 

In the maximum-entropy method, 
we use the same four types of morphological information, 
(i) major POS, (ii) minor POS, (iii) semantic information, and 
(iv) word, as in the decision-tree method. 
However, 
it does not consider 
a combination of features. 
Unlike the decision-tree method, 
as a result 
we had to combine features manually. 

First we considered a combination of the bits of 
each morphological information. 
Because there were four types of information, 
the total number of combinations was $2^{4}-1$. 
Since this number is large and intractable, 
we considered that 
(i) major POS, (ii) minor POS, (iii) semantic information, and 
(iv) word information gradually become more specific in this order, 
and we combined the four types of information 
in the following way: 
\begin{equation}
\small
  \label{equ:info}
  \begin{minipage}[h]{14cm}
    \begin{tabular}[h]{l@{ }l}
Information A: & (i) major POS\\
Information B: & (i) major POS and (ii) minor POS\\
Information C: & (i) major POS, (ii) minor POS and\\
& (iii) semantic information\\
Information D: & (i) major POS, (ii) minor POS, \\
& (iii) semantic information and (iv) word \\
    \end{tabular}
  \end{minipage}
\end{equation}
We used only Information A and B 
for the two outside morphemes  
because we did not use semantic and word information 
in the same way it is used in the decision-tree method. 

\begin{figure*}[t]
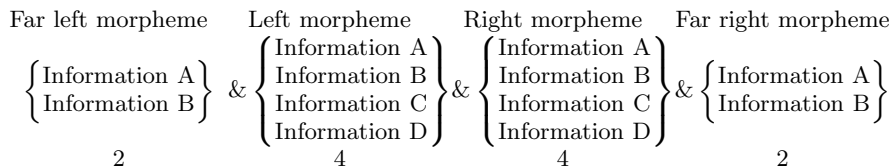

\small
  \begin{center}
{
\[
\begin{array}[h]{c@{}c@{}c@{}c}
\mbox{Far left morpheme \, } & \mbox{Left morpheme \, }  & \mbox{Right morpheme \, } & \mbox{Far right morpheme}\\
\left\{
\begin{array}[h]{@{}c@{}}
\mbox{Information A}  \\
\mbox{Information B}  
\end{array}
\right\} 
&
\&\left\{
\begin{array}[h]{@{}c@{}}
\mbox{Information A}  \\
\mbox{Information B}  \\
\mbox{Information C}  \\
\mbox{Information D}  
\end{array}
\right\} 
&
\&\left\{
\begin{array}[h]{@{}c@{}}
\mbox{Information A}  \\
\mbox{Information B}  \\
\mbox{Information C}  \\
\mbox{Information D}  
\end{array}
\right\} 
&
\&\left\{
\begin{array}[h]{@{}c@{}}
\mbox{Information A}  \\
\mbox{Information B}  
\end{array}
\right\} \\
2 & 4 & 4 & 2
\end{array}
\]
}
\vspace*{-0.5cm}
    \caption{Features used in the maximum-entropy method.}
    \label{fig:info_me}
  \end{center}
\end{figure*}

Next, we considered the combinations of each type of information. 
As shown in Figure \ref{fig:info_me}, 
the number of combinations was 
64 ($2 \times 4 \times 4 \times 2$). 

For data sparseness, 
in addition to the above combinations,  
we considered the cases in which first,  
one of the two outside morphemes was not used, 
secondly, 
neither of the two outside ones were used, 
and thirdly,  
only one of the two middle ones is used. 
The number of features used in the maximum-entropy method 
is 152, which is obtained 
as follows:\footnote{
When we extracted features from 
all of the articles on January 1, 1995 of 
a Kyoto University corpus 
(the number of spaces between morphemes was 25,814) 
by using this method, 
the number of types of features was 1,534,701. }

{
\begin{tabular}[h]{c@{}p{0.5cm}@{}p{0.5cm}@{}p{0.5cm}@{}p{0.5cm}@{}p{0.5cm}@{}p{0.5cm}@{}p{0.5cm}@{}p{0.5cm}@{}p{0.5cm}@{}}
  No. of features & $=$ & 2 & $\times$  &4  &$\times$  &4  &$\times$  &2 \\
           & $+$ & 2 & $\times$  &4 & $\times$  &4      \\
           & $+$ &   &     &4  &$\times$  &4  &$\times$  &2             \\
           & $+$ &   &     &4  &$\times$  &4 \\
           & $+$ &   &     &4 & \\
           & $+$ &   &      &  &    &4 \\
           & $=$ & 152 &\\
\end{tabular}
}

In Figure \ref{fig:kaiseki_info}, 
the feature that uses 
Information B in the far left morpheme, 
Information D in the left morpheme, 
Information C in the right morpheme, and 
Information A in the far right morpheme is 
``Noun: Normal Noun; Particle: Case-Particle: none: {\it wo}; Verb: Normal Form: 217; Symbol''. 
In the maximum-entropy method 
we used for each space 152 features such as this one. 

\subsection{Example-based method (use of similarity)}

An example-based method was proposed by 
Nagao \cite{Nagao_EBMT} 
in an attempt 
to solve problems in machine translation. 
To resolve a problem, 
it uses the most similar example. 
In the present work, the example-based method impartially used 
the same four types of information (see Eq. (1)) 
as used in the maximum-entropy method. 

To use this method, 
we must define the similarity of an input to an example. 
We use the 152 patterns from the maximum-entropy method 
to establish the level of similarity. 
We define the similarity $S$ between an input and an example 
according to which one of these 152 levels is the matching level, 
as follows. 
(The equation reflects the importance of the two middle morphemes.)\\
\begin{equation}
  \label{eqn:ruijido}
  \begin{minipage}[h]{6.5cm}
S $=$ s($m_{-1}$) $\times$ s($m_{+1}$) $\times$ 10,000

\hspace*{0.2cm} $+$ s($m_{-2}$) $\times$ s($m_{+2}$)
\end{minipage}
\end{equation}\\
Here $m_{-1}$, $m_{+1}$, $m_{-2}$, and $m_{+2}$ refer respectively 
to the left, right, far left, 
and far right morphemes, 
and s(x) is the morphological similarity of a morpheme x, 
which is defined as follows: \\
\begin{equation}
\small
  \label{eqn:ruijido2}
  \begin{minipage}[h]{7.5cm}
\begin{tabular}[h]{l@{}l}
s(x) $=$&1 (when no information of x is matched) \\
       &2 (when Information A of x is matched)\\
       &3 (when Information B of x is matched)\\
       &4 (when Information C of x is matched)\\
       &5 (when Information D of x is matched)\\
\end{tabular}
  \end{minipage}
\end{equation}

\begin{figure*}[t]
\small
  \begin{center}
\begin{tabular}[h]{|lcllll|}\hline
     &        & {\it bun} & {\it wo} & {\it kugiru} & . \\
     &        & (sentence) & ({\sf obj}) & (divide) & . \\
     & s(x)& $m_{-2}$&$m_{-1}$&$m_{+1}$&$m_{+2}$\\
No information& 1 & --- & --- & --- & ---\\
Information A & 2& Noun & Particle & Verb & Symbol\\
Information B & 3 & Normal Noun & Case-Particle & Normal Form & Punctuation\\
Information C & 4 & $\times$ & None & 217 & $\times$\\
Information D & 5 & $\times$ & {\it wo} & {\it kugiru} & $\times$\\\hline
\end{tabular}
    \caption{Example of levels of similarity}
    \label{fig:ruijido_rei}
  \end{center}
\end{figure*}

Figure \ref{fig:ruijido_rei} shows an example of the levels of similarity. 
When a pattern matches 
Information A of all four morphemes, 
such as ``Noun; Particle; Verb; Symbol'', 
its similarity is 40,004 ($2 \times 2 \times 10,000 + 2 \times 2$). 
When a pattern matches a pattern, 
such as ``---; Particle: Case-Particle: none: {\it wo}; ---; ---'', 
its similarity is 50,001 ($5 \times 1 \times 10,000 + 1 \times 1$). 

The example-based method extracts 
the example with the highest level of similarity and 
checks whether or not that example is marked.   
A partition mark is inserted in the input data 
only when the example is marked. 
When multiple examples have the same highest level of similarity, 
the selection of the best example is ambiguous. 
In this case, 
we count the number of marked and 
unmarked spaces in all of the examples and 
choose the larger.

\subsection{Decision-list method (use of probability and frequency)}

The decision-list method was 
proposed by Rivest \cite{Rivest:ML87},  
in which the rules are not expressed as a tree structure 
like in the decision-tree method, 
but are expanded by combining all the features,  
and are stored in a one-dimensional list. 
A priority order is defined in a certain way and 
all of the rules are arranged in this order. 
The decision-list method searches for rules 
from the top of the list 
and analyzes a particular problem by using only the first applicable rule. 

In this study we used 
in the decision-list method 
the same 152 types of patterns 
that were used in the maximum-entropy method. 

To determine the priority order of the rules, 
we referred to Yarowsky's method \cite{Yarowsky:ACL94} and Nishiokayama's method \cite{Nishiokayama98e} 
and used the probability and frequency of each rule as measures 
of this priority order. 
When multiple rules had the same probability, 
the rules were arranged in order of their frequency. 

Suppose, for example, that 
Pattern A ``Noun: Normal Noun; Particle: Case-Particle: none: {\it wo}; Verb: Normal Form: 217; Symbol: Punctuation'' occurs 13 times in a learning set 
and that ten of the occurrences include the inserted partition mark. 
Suppose also that 
Pattern B ``Noun; Particle; Verb; Symbol'' occurs 123 times
in a learning set and that 90 of the occurrences include the mark. 

This example is recognized by the following rules: 

\begin{quote}
\small
\mbox{Pattern A $\Rightarrow$ Partition  \, 76.9\% (10/ 13), \, Freq. 23 }

\mbox{Pattern B $\Rightarrow$ Partition \, 73.2\% (90/123), \, Freq. 123}
\end{quote}

Many similar rules were made and were 
then listed in order of their probabilities and, 
for any one probability, in order of their frequencies. 
This list was searched from the top 
and the answer was obtained by using the first applicable rule. 

\subsection{Method 1 (use of category-exclusive rules)}
\label{sec:haiha}
\label{sec:new1}

So far, we have 
described the four existing machine learning methods. 
In the next two sections we describe our methods. 

It is reasonable to consider 
the 152 patterns used in three of the previous methods. 
Now, let us suppose that 
the 152 patterns from the learning set 
yield the statistics of Figure \ref{fig:rule_method1}. 

\begin{figure*}[t]
\small
  \begin{center}
{
\begin{tabular}[h]{llllr@{ }cl}
Rule A: & Pattern A & $\Rightarrow$ & probability of non-partition &100\% & ( 34/ 34) & Frequency 34\\
Rule B: & Pattern B & $\Rightarrow$ & probability of partition &100\% & ( 33/ 33)& Frequency 33 \\
Rule C: &Pattern C & $\Rightarrow$ & probability of partition &100\% & ( 25/ 25) & Frequency 25 \\
Rule D: &Pattern D & $\Rightarrow$ & probability of partition &100\% & ( 19/ 19) & Frequency 19 \\
Rule E: &Pattern E & $\Rightarrow$ & probability of partition &81.3\% &(100/123) & Frequency 123\\
Rule F: &Pattern F & $\Rightarrow$ & probability of partition &76.9\% &( 10/ 13) & Frequency 13\\
Rule G: &Pattern G & $\Rightarrow$ & probability of non-partition &57.4\% & (310/540)& Frequency 540 \\
.....  & ..... & .....\\
\end{tabular}}
    \caption{an example of rules used in Method 1}
    \label{fig:rule_method1}
  \end{center}
\end{figure*}

``Partition'' means that the rule determines that 
a partition mark should be inserted in the input data 
and ``non-partition'' means that the rule determines that 
a partition mark should not be inserted. 

Suppose that 
when we solve a hypothetical problem 
Patterns A to G are applicable. 
If we use the decision-list method, 
only Rule A is used, which is applied first, 
and this determines that a partition mark should not be 
inserted. 
For Rules B, C, and D, 
although the frequency of each rule is lower than that of Rule A, 
the sum of their frequencies of the rules is higher, 
so we think that 
it is better to use Rules B, C, and D than Rule A. 
Method 1 follows this idea, but 
we do not simply sum up the frequencies. 
Instead, we count the number of examples used in Rules B, C, and D 
and judge 
the category having the largest number of examples 
that satisfy the pattern with the highest probability 
to be the desired answer. 

For example, suppose that in the above example 
the number of examples satisfying Rules B, C, and D is 65. 
(Because some examples overlap in multiple rules, 
the total number of examples is actually smaller than 
the total number of the frequencies of the three rules.) 
In this case, 
among the examples used by the rules having 100\% probability, 
the number of examples of partition is 65, 
and the number of examples of non-partition is 34. 
So, we determine that the desired answer is to partition. 

A rule having 100\% probability is called 
{\it a category-exclusive rule} 
because 
all the data satisfying it belong to 
one category, which is either partition or non-partition. 
Because for any given space 
the number of rules used can be as large as 152, 
category-exclusive rules are applied often\footnote{
The ratio of the spaces analyzed by using 
category-exclusive rules is 99.30\% (16864/16983) 
in Experiment 1 of Section \ref{sec:jikken}. 
This indicates that 
almost all of the spaces are analyzed by category-exclusive rules.}. 
Method 1 uses all of these category-exclusive rules, 
so we call it the {\it method using category-exclusive rules}.

Solving problems by using rules whose probabilities are 
not 100\% may result in the wrong solutions. 
Almost all of the traditional machine learning methods 
solve problems by using rules whose probabilities are not 100\%. 
By using such methods, 
we cannot hope to improve accuracy. 
If we want to improve accuracy, 
we must use category-exclusive rules. 
There are some cases, however, for which, 
even if we take this approach, 
category-exclusive rules are rarely applied. 
In such cases, 
we must add new features to the analysis 
to create a situation in which 
many category-exclusive rules can be applied. 

However, it is not sufficient to use category-exclusive rules. 
There are many meaningless rules 
which happen to be category-exclusive only in a learning set. 
We must consider how to eliminate such meaningless rules. 

\subsection{Method 2 (using category-exclusive rules 
with the highest similarity)}
\label{sec:new2}

Method 2 combines the example-based method and Method 1. 
That is, it combines 
the method using similarity and 
the method using category-exclusive rules
in order to eliminate 
the meaningless category-exclusive rules 
mentioned in the previous section. 

Method 2 also uses 152 patterns for identifying bunsetsu. 
These patterns are used as rules in the same way as in Method 1. 
Desired answers are determined 
by using the rule having the highest probability. 
When multiple rules have the same probability, 
Method 2 uses the value of the similarity described 
in the section of the example-based method 
and analyzes the problem with the rule having the highest similarity. 
When multiple rules have the same probability and similarity, 
the method takes the examples used by the rules 
having the highest probability and the highest similarity, 
and chooses the category with the larger number of examples 
as the desired answer, 
in the same way as in Method 1. 

However, 
when category-exclusive rules having more than one frequency exist, 
the above procedure is performed 
after eliminating 
all of the category-exclusive rules having one frequency. 
In other words, 
category-exclusive rules having more than one frequency 
are given a higher priority than 
category-exclusive rules having only one frequency 
but having a higher similarity. 
This is because 
category-exclusive rules having only one frequency 
are not so reliable.

\section{Experiments and discussion}
\label{sec:jikken}

In our experiments 
we used a Kyoto University text corpus \cite{kurohashi:nlp97_e}, 
which is a tagged corpus made up of articles from the Mainichi newspaper.  
All experiments reported in this paper were performed using 
articles dated from January 1 to 5, 1995. 
We obtained the correct information 
on morphology and bunsetsu identification from the tagged corpus. 

The following experiments 
were conducted to determine which supervised learning method 
achieves the highest accuracy rate. 
\begin{itemize}
\item
  Experiment 1

  Learning set: January 1, 1995

  Test set: January 3, 1995

\item 
  Experiment 2

  Learning set: January 4, 1995

  Test set: January 5, 1995

\end{itemize}

Because we used Experiment 1 
in making Method 1 and Method 2, 
Experiment 1 is a closed data set for Method 1 and Method 2. 
So, we performed Experiment 2. 

The results are listed in 
Tables \ref{tab:jikken1_close} to \ref{tab:jikken2_open}. 
We used KNP2.0b4 \cite{KNP2.0b4_e} and KNP2.0b6 \cite{KNP2.0b6_e}, 
which are bunsetsu identification and syntactic analysis systems 
using many hand-made rules 
in addition to the six methods described in Section \ref{sec:shuhou}. 
Because KNP is not based on a machine learning method but 
many hand-made rules, 
in the KNP results ``Learning set'' and ``Test set'' 
in the tables have no meanings. 
In the experiment of KNP, 
we also uses morphological information in a corpus. 
The ``F'' in the tables indicates the F-measure, 
which is the harmonic mean of a recall and a precision. 
A recall is the fraction of correctly identified partitions 
out of all the partitions. 
A precision is 
the fraction of correctly identified partitions 
out of all the spaces which were judged to have 
a partition mark inserted. 
\Small{
We used the system made by Ristad \cite{ristad98} in the 
maximum-entropy experiment, 
but the system did not work 
with features having low frequency. 
After eliminating 
all the combinations\footnote{
\baselineskip=0.9\baselineskip
The frequency of the combination 
of ``{\it a certain feature}'' and 
``{\it the category when the feature exists}'' 
is necessary as the input of the system. 
The system worked 
when we eliminated combinations having low frequency. 
So, we conducted experiments in this way. 
However, a better way may exist. 
More features are eliminated when we eliminate combinations 
of a feature and a category 
than when we eliminate features by their frequencies. 
}
 of a feature and category 
having a frequency of less than 12, 
we ran the system again.
(The system did not work 
when we eliminated combinations with a frequency of less than 11.) 
Therefore, 
it may be possible to obtain a higher accuracy rate 
with the maximum entropy method than here. 
In the decision-tree method, 
we used C4.5 \cite{c4.5}. 
In the experiments we used the -s option, 
which groups feature values. 
However, this grouping takes a long time when the 
number of feature values is large. 
So, we carried out experiments 
after preparing feature values 
called ``the others'' and changing 
the feature values having a frequency of less than 10 
into ``the others'' feature values. 
}

\begin{table}[t]
\small
\baselineskip=0.85\baselineskip
\caption{Results of learning set of Experiment 1}
    \label{tab:jikken1_close}
  \begin{center}
  \begin{tabular}[h]{|l|r|r|r|}\hline
\multicolumn{1}{|c|}{Method}      & \multicolumn{1}{|c|}{F}   & \multicolumn{1}{|c|}{Recall} & \multicolumn{1}{|c|}{Precision}\\\hline
Decision Tree    &  99.58\% &  99.66\% &  99.51\% \\
Maximum Entropy  &  99.20\% &  99.35\% &  99.06\% \\
Example-Based &  99.98\% & 100.00\% &  99.97\% \\
Decision List&  99.98\% & 100.00\% &  99.97\% \\
Method 1  &  99.98\% & 100.00\% &  99.97\% \\
Method 2   &  99.98\% & 100.00\% &  99.97\% \\\hline
KNP 2.0b4 &  99.23\% &  99.78\% &  98.69\% \\
KNP 2.0b6 &  99.73\% &  99.77\% &  99.69\% \\\hline
\end{tabular}\\
The number of spaces between two morphemes is 25,814. 
The number of partitions is 9,523. 
\end{center}
\end{table}

\begin{table}[t]
\small
\baselineskip=0.85\baselineskip
    \caption{Results of test set of Experiment 1}
    \label{tab:jikken1_open}
  \begin{center}
  \begin{tabular}[h]{|l|r|r|r|}\hline
\multicolumn{1}{|c|}{Method}      & \multicolumn{1}{|c|}{F}   & \multicolumn{1}{|c|}{Recall} & \multicolumn{1}{|c|}{Precision}\\\hline
Decision Tree    &  98.87\% &  98.67\% &  99.08\% \\
Maximum Entropy  &  98.90\% &  98.75\% &  99.06\% \\
Example-Based &  99.02\% &  98.69\% &  99.36\% \\
Decision List&  98.95\% &  98.43\% &  99.48\% \\
Method 1   &  98.98\% &  98.54\% &  99.43\% \\
Method 2   &  99.16\% &  98.88\% &  99.45\% \\\hline
KNP 2.0b4 &  99.13\% &  99.72\% &  98.54\% \\
KNP 2.0b6 &  99.66\% &  99.68\% &  99.64\% \\\hline
\end{tabular}\\
The number of spaces between two morphemes is 16,983. 
The number of partitions is 6,166. 
\end{center}
\end{table}

\begin{table}[t]
\small
\baselineskip=0.85\baselineskip
    \caption{Results of learning set of Experiment 2}
    \label{tab:jikken2_close}
  \begin{center}
  \begin{tabular}[h]{|l|r|r|r|}\hline
\multicolumn{1}{|c|}{Method}      & \multicolumn{1}{|c|}{F}   & \multicolumn{1}{|c|}{Recall} & \multicolumn{1}{|c|}{Precision}\\\hline
Decision Tree   &  99.70\% &  99.71\% &  99.69\% \\
Maximum Entropy  &  99.07\% &  99.23\% &  98.92\% \\
Example-Based &  99.99\% & 100.00\% &  99.98\% \\
Decision List &  99.99\% & 100.00\% &  99.98\% \\
Method 1  &  99.99\% & 100.00\% &  99.98\% \\
Method 2   &  99.99\% & 100.00\% &  99.98\% \\\hline
KNP 2.0b4 &  98.94\% &  99.50\% &  98.39\% \\
KNP 2.0b6 &  99.47\% &  99.47\% &  99.48\% \\\hline
\end{tabular}\\
The number of spaces between two morphemes is 27,665. 
The number of partitions is 10,143. 
\end{center}
\end{table}

\begin{table}[t]
\small
\baselineskip=0.85\baselineskip
    \caption{Results of test set of Experiment 2}
    \label{tab:jikken2_open}
  \begin{center}
  \begin{tabular}[h]{|l|r|r|r|}\hline
\multicolumn{1}{|c|}{Method}      & \multicolumn{1}{|c|}{F}   & \multicolumn{1}{|c|}{Recall} & \multicolumn{1}{|c|}{Precision}\\\hline
Decision Tree  &  98.50\% & 98.51\% & 98.49\%\\
Maximum Entropy &  98.57\% & 98.55\% & 98.59\%\\
Example-Based &  98.82\% & 98.71\% & 98.93\%\\
Decision List&  98.75\% & 98.27\% & 99.23\%\\
Method 1   &  98.79\% & 98.54\% & 99.43\%\\
Method 2   &  98.90\% & 98.65\% & 99.15\%\\\hline
KNP 2.0b4 &  99.07\% & 99.43\% & 98.71\%\\
KNP 2.0b6 &  99.51\% & 99.40\% & 99.61\%\\\hline
\end{tabular}\\
The number of spaces between two morphemes is 32,304. 
The number of partitions is 11,756. 
\end{center}
\end{table}

Tables \ref{tab:jikken1_close} to \ref{tab:jikken2_open} 
show the following results: 
\begin{itemize}
\item 
  In the test set 
  the decision-tree method was a little better than 
  the maximum-entropy method. 
\Small{
\footnote{
    \baselineskip=0.9\baselineskip
    It is not a perfect comparison 
    because we eliminated features in the experiments of both methods. 
    However, because we eliminated more features 
    in the maximum entropy method 
    than in the decision-tree method, 
    we can expect that maximum entropy method 
    having a little better accuracy rate 
    is better than decision tree method. 
}. }%
Although the maximum-entropy method has 
a weak point in that it does not learn the combinations of features, 
we could overcome this weakness 
by making almost all of the combinations of features 
to produce a higher accuracy rate. 
\item 
  The decision-list method was better than 
  the maximum-entropy method in this experiment. 
  \Small{
    However, we had to eliminate some of the features 
  for the maximum-entropy method 
  because of the system problem. 
  The accuracy rate may be higher 
  if we can use these eliminated features. 
  }
\item 
  The example-based method obtained the highest accuracy rate 
  among the four existing methods. 
\item 
  Although Method 1, which uses the category-exclusive rule, was 
  worse than the example-based method, 
  it was better than the decision-list method. 
  One reason for this was that the decision-list method 
  chooses rules randomly when 
  multiple rules have identical probabilities and frequencies. 
\item 
  Method 2, 
  which uses the category-exclusive rule with the highest similarity, 
  achieved the highest accuracy rate 
  among the supervised learning methods. 
\item 
  The example-based method, the decision-list method, 
  Method 1 and Method 2 obtained accuracy rates of 
  about 100\% for the learning set. 
  This indicates that these methods are 
  especially strong for learning sets. 
\item 
  The two methods using similarity (example-based method and 
  Method 2) were always better than the other methods, 
  indicating that 
  the use of similarity is effective 
  if we can define it appropriately. 
\item 
  We carried out experiments by using KNP, 
  a system that uses many hand-made rules. 
  The F-measure of KNP was highest in the test set. 
\item 
  We used two versions of KNP, KNP 2.0b4 and KNP 2.0b6. 
  The latter was much better than the former, 
  indicating that 
  the improvements made by hand are effective. 
  But, the maintenance of rules by hand has a limit, 
  so the improvements made by hand are not always effective. 

\end{itemize}

The above experiments indicate that 
Method 2 is best among the machine learning methods\footnote{In 
these experiments, the differences were very small. 
But, we think that 
the differences are significant to some extent 
because we performed 
Experiment 1 and Experiment 2, 
the data we used are a large corpus 
containing about a few ten thousand morphemes 
and tagged objectively in advance, 
and the difference of about 0.1\% is large 
in the precisions of 99\%.}. 

In Table \ref{tab:seikairei} we show some cases 
which were partitioned incorrectly with KNP but correctly with Method 2. 
A partition with ``NEED'' indicates 
that KNP missed inserting the partition mark, and 
a partition with ``WRONG'' indicates 
that KNP inserted the partition mark incorrectly. 
In the test set of Experiment 1, 
the F-measure of KNP2.0b6 was 99.66\%. 
The F-measure increases to 99.83\%, 
under the assumption that 
when KNP2.0b6 or Method 2 is correct, the answer is correct. 
Although the accuracy rate for KNP2.0b6 was high, 
there were some cases in which KNP partitioned incorrectly 
and Method 2 partitioned correctly.  
A combination of Method 2 with KNP2.0b6 may be able to improve the F-measure. 

\begin{table}[t]
\small
\baselineskip=0.85\baselineskip
\caption{Cases when KNP was incorrect and Method 2 was correct}
\label{tab:seikairei}
\vspace*{0.5cm}
\begin{tabular}{|l@{ }l@{ }l|}\hline
{\it kotsukotsu} & $|^{\tiny NEED}$ &{\it gaman-shi} \\
(steadily) & &  (be patient with) \\
\multicolumn{3}{|l|}{(... be patient with ... steadily)}\\\hline
\end{tabular}
\begin{tabular}{|@{ }l@{ }l@{ }l@{ }l@{ }l@{ }l@{ }|}\hline
{\it yoyuu} & {\it wo} & $|$ & {\it motte} & $|^{\tiny NEED}$ & {\it shirizoke}\\
(enough strength) & {\sf obj} & & (have) &  & (beat off)\\
\multicolumn{6}{|l|}{(... beat off ... having enough strength)}\\\hline
\end{tabular}
\begin{tabular}{|@{ }l@{ }l@{ }l@{ }l@{ }l@{ }l@{ }|}\hline
{\it kaisha} & {\it wo} & $|$ & {\it gurupu-wake} & $|^{\tiny WRONG}$ & {\it shite}  \\
{\it company} & {\sf obj} & & (grouping) &  & (do) \\
\multicolumn{6}{|l|}{(... do grouping companies)}\\\hline
\end{tabular}
\None{
\begin{tabular}{|lllll|}\hline
{\it mottomo} & $|$ & {\it nare} & $|^{\tiny WRONG}$ & {\it shitashinde-iru}\\
(best) & & (get used to (\sf prefix)) & & (be familiar with)\\
\multicolumn{5}{|l|}{(... be familiar with ... best)}\\\hline
\end{tabular}}
\end{table}

The only previous research 
resolving bunsetsu identification by machine learning methods, 
is the work by Zhang \cite{Zhang98}. 
The decision-tree method was used in this work. 
But this work used only a small number of information 
for bunsetsu identification\footnote{This work used 
only the POS information of the two morphemes of an analyzed space. 
} and did not achieve high accuracy rates. 
(The recall rate was 97.6\%(=2502/(2502+62)), 
the precision rate was 92.4\%(=2502/(2502+205)), 
and F-measure was 94.2\%.)

\section{Conclusion}

To solve the problem of accurate bunsetsu identification, 
we carried out experiments 
comparing four existing machine-learning methods 
(decision-tree method, maximum-entropy method, example-based method and 
decision-list method). 
We obtained the following order of accuracy in bunsetsu identification. 

\begin{center}
  Example-Based $>$ Decision List $>$ 

  Maximum Entropy $>$ Decision Tree
\end{center}

We also described a new method which uses 
category-exclusive rules with the highest similarity. 
This method performed better than the other learning methods 
in our experiments. 

{\footnotesize

\end{document}